\documentclass{article}
\usepackage{natbib}
\usepackage[utf8]{inputenc}
\usepackage{amsmath,amssymb}
\usepackage{graphicx}
\usepackage[utf8]{inputenc}
\usepackage[english]{babel}
\usepackage[toc,page]{appendix}
\usepackage[colorlinks=TRUE, linkcolor=blue, urlcolor=blue, citecolor=blue]{hyperref}
\usepackage[paper=portrait,pagesize]{typearea}
\graphicspath{ {./images/} }

\usepackage{natbib}
\usepackage[nocenter]{qtree}
\usepackage{tree-dvips}
\usepackage{tikz}
\usetikzlibrary{arrows}
\usepackage{breqn}
\usepackage{multirow}
\usepackage{floatrow}
\usepackage{authblk}
\usepackage{float}
\usepackage[toc,page]{appendix}

\usepackage{tikz}
\usetikzlibrary{bayesnet}

\tikzset{
  treenode/.style = {align=center, inner sep=0pt, text centered,
    font=\sffamily},
  arn_n/.style = {treenode, circle, black, font=\sffamily\bfseries, draw=black,
    fill=white, text width=1.5em}
}
\usepackage{tikz}

\title{Drift Estimation with Graphical Models}
\author[1,*]{Luigi Riso}
\author[2]{Marco Guerzoni}
\affil[1]{University of Turin}
\affil[2]{DEMS, University of Milan-Bicocca}
\date{January 2020}

\begin{document}

\maketitle
\let\oldthefootnote\thefootnote
\renewcommand{\thefootnote}{\fnsymbol{footnote}}
\footnotetext[1]{To whom correspondence should be addressed.  Department of Economics and Statistics, Lungo Dora Siena 100A, 10122, Turin, Italy. Email: \url{luigi.riso@unito.it}}
\let\thefootnote\oldthefootnote

\begin{abstract}

This paper deals with the issue of concept drift in supervised machine learning. We make use of graphical models to elicit the visible structure of the data and we infer from there changes in the hidden context. Differently from previous concept-drift-detection methods, this application does not depend on the supervised machine learning model in use for a specific target variable, but it tries to assess the concept drift as independent characteristic of the evolution of a data set. Specifically we investigate how a graphical model evolves by looking at the creation of new links and the disappearing of existing ones in different time periods. The paper suggests a method that highlights the changes and eventually produce a metric to evaluate the stability over time. The paper evaluate the method with real world data on the Australian Electric market.
\end{abstract}

\section{Introduction}
 In the last decades, both the increasing availability of digitised information and the improvement in the algorithms made the use of machine learning widespread across different industries. Specifically, supervised machine learning became a standard tool for predicting key information in various organization processes such as for instance to mention a few risk default of firms and individual, fraudulent claims, customers churn, and machine failures. The assessment of model uncertainty within a supervised machine learning exercise is based on testing the goodness on a test-set, whose observations have not been employed in the model training.  This practice allows for flexibility in the choice of the model and  prevents from the risk of over-fitting. 
 However, this analysis relies on the assumption the data generating structure is similar between the test-set and the future observations. While this assumption is rarely debatable in physical process, social process change overtime and a model trained on past data might see a deterioration of its predictive power \cite{gama2014survey}. This phenomenon is known as concept- or model- drift and describes the situation in which there exists an hidden context of data generative structure, that is any effect of the outcome variable not captured by the model features, which changes over time abruptly, incrementally, or periodically \cite{widmer1996learning, webb2016characterizing}.
 Scholars addressed this issue and developed a battery of techniques for concept drift detection and early detection. 
 As reviewed in \citet{klinkenberg2000detecting} and \citet{elwell2011incremental}, traditional techniques in concept drift detection typically relies by adopting different time windows or size of the training data \citep{klinkenberg1998adaptive} or in explaining how the weights of different features change overtime in the outcome prediction \citep{klinkenberg1998adaptive, taylor1997structural, klinkenberg2004learning}.
A recent review \citep{althabiti2020classification} surveys also methods which can also deal with model update with stream data \citep{bose2011handling}. However, all of these techniques rely on some sort of computation or statistical comparison of the changes on classification error overtime and from this evidence they deduct  the presence of concept drift \citep{widmer1996learning}. In this paper, we approach the problem from a different angle. We make use of graphical models \citep{lauritzen1996graphical} to elicit the visible structure of the data and we infer from there changes in the hidden context with use of statistical measure. Thus, differently from previous concept-drift-detection methods, this application does not depend on the supervised machine learning model in use, but it tries to assess the concept drift as an independent characteristic of the evolution of a data set.

\section{Graphical Models, background}
Consider a dataset, composed by \textit{p} random variables $\textbf{X}_p$, where $p$ can be divided in \textit{d} discrete and \textit{q} continuous random variables. Graphical Models are a method to display the conditional independence relationships between random variables in a dataset. The conditional independence relationships can be showed as a networks of variables with an undirected  graph, that is 
 mathematical object \begin{math}
G=(V, E)
\end{math}, where V is a finite set of nodes, one-to-one correspondence with the $p$ random variables present in the dataset, and \begin{math}
E \subset V \times V 
\end{math}, is a subset of ordered couples of V.
Links represent interactions between the nodes. If a link between two nodes is absent, the two variables represented by the node are conditional independent given the dependence of the remaining variables. 

Pairwise, local and global Markov properties are the connections between graph theory and statistical modeling \citep{lauritzen1996graphical}.  As said before, there exist a one-to-one correspondence between the variables and the nodes in the graph and, for this reason, the sets of nodes is \begin{math}
\Delta
\end{math} and \begin{math}
\Gamma
\end{math}, where \begin{math}
V=\{ \Delta \cup \Gamma \}
\end{math}. Let the corresponding random variables be \begin{math}
(\textbf{Z},\textbf{Y})
\end{math} where \begin{math} \textbf{Z}=(\textit{Z}_1,...,\textit{Z}_d)
\end{math} and \begin{math}
\textbf{Y}=(\textit{Y}_1,...,\textit{Y}_q)
\end{math} and a \textit{i}-observation be \begin{math}
(\textbf{z}_i,\textbf{y}_i)
\end{math}. This means that \begin{math}
\textbf{z}
\end{math} is a d-tuple containing the values of discrete variables, and  \begin{math}
\textbf{y}
\end{math} is a real vector of length \textit{q}. Our interest is to estimate the joint probability distribution \begin{math}
P(\textbf{x})
\end{math}
for the random variables \begin{math}
(\textbf{Z}, \textbf{Y})
\end{math}  to build a conditional (undirected)  graph from the data. A product approximation of \begin{math}
P(\textbf{x})
\end{math} is defined to be a product of several of its component distribution of lower order  \begin{math}
P_a(\textbf{x})
\end{math}.
As suggest \cite{chow1968approximating}, we can consider the class of second-order approximation, i.e: 
\begin{equation}
    P_a(\textbf{x})=\prod_{i=1}^{p} P(x_i,x_{j(i)}), \quad
    0\leqslant j(i) \leqslant p
\end{equation}
where \begin{math}
(j_1,...,j_p)
\end{math} is an unknown permutation of integers \begin{math}
(1,2,...,p)
\end{math}, where \begin{math}
\textit{p=d+q}
\end{math}.
Chow and Liu  in [6] proved that for discrete random \begin{math}
\textbf{Z}
\end{math}, the problem of finding the goodness of approximation between \begin{math}
P(\textbf{z})
\end{math} and \begin{math}
P_a(\textbf{z})
\end{math} with the minimization of the closeness measure:
\begin{equation}
    \textit{I}(P, P_a)=\sum_{z}P(\textbf{z}) \log \frac{P(\textbf{z})}{P_a(\textbf{z})}
\end{equation}
where \begin{math}
\sum_z P(\textbf{z})
\end{math}  is nothing more than the sum over all levels of discrete variables. The equation (2), is equivalent to maximizing the total branch (link) weight \begin{math}
\sum_{i=1}^{p} \textit{I}(z_i, z_{j(i)})
\end{math}, where:
\begin{equation}
    \textit{I}(z_i, z_{j(i)})=\sum_{z_i,z_{j(i)}}P(z_i,z_{j(i)}) \log \left(\frac{P(z_i,z_{j(i)})}{P(z_i)P(z_{j(i)})} \right)
\end{equation}
The task is to build a tree or forest (different trees) of maximum weight. We make use of the Kruskal's algorithm \citep{kruskal1956shortest} to  compute trees with the minimum of total length. To choose a tree of maximum total branch weight, we first index the \begin{math}
\textit{d}(\textit{d}-1)/2
\end{math} according to decreasing weight. This algorithm starts from a square weighted matrix \begin{math} d \times d
\end{math}, where a weight for a couple of variables \begin{math}
(Z_i,Z_j)
\end{math} is given by  the mutual information \begin{math}
\textit{I}(z_i,z_j)
\end{math}. In the real world the probability distributions are no given explicitly, for this reason we have to estimate the mutual information. Let \begin{math}
\textbf{z}^1,\textbf{z}^2,...,\textbf{z}^N
\end{math} be independent samples of finite discrete variables \begin{math}
\textbf{z}
\end{math}. Then the mutual information is given by:
\begin{equation}
    \hat{I}(z_i,z_j)=\sum_{u,v}f_{u,v}(i,j) \log \frac{f_{u,v}(i,j)}{f_u(i) f_v(j)},
\end{equation}
where \begin{math}
f_{u,v}(i,j)=\frac{n_{uv}(i,j)}{\sum_{u v} n_{uv}(i,j)}
\end{math} and \begin{math}
n_{u v}(i,j)
\end{math} is the number of samples such that their \textit{i}th and \textit{j}th components assume the values of \textit{u} and \textit{v}, respectively. It was showed that with this estimator we also maximize the likelihood for a dependence tree \cite{chow1968approximating}. This procedure works only with the discrete random variables, but it can be extended to data with both discrete and continuous random variables \cite{edwards2010selecting}.
To present this extension, we have to consider the distributional assumption of our random variables \begin{math}
\textbf{X}
\end{math} i.e. the distribution of \begin{math}
\textbf{Y}
\end{math} given \begin{math}
\textbf{Z}=\textbf{z}
\end{math} is a multivariate normal \begin{math}
\mathcal{N}(\mu_i, \Sigma_i)
\end{math} so that both the conditional mean and covariance may depend on \textit{i}th component. \\
We distinguish between homogenous and heterogeneous case, if \begin{math}
\Sigma 
\end{math} depend on \textit{i} we are in the homogenous case, otherwise we are in the heterogeneous case. More details this conditional Gaussian distribution can be found in \cite{sudderth2004embedded}. Before to apply the Kruskal's algorithm, we need to find an estimator of the mutual information \begin{math}
I(z_u,y_v)
\end{math} between each couple of variables in the mixed case. For a couple of variables \begin{math}
(Z_u, Y_v)
\end{math} we can write the sample cell count, mean, and finally  the variance, respectively, \begin{math}
\{ n_i, \bar{y}_v, s_i^{(v)}\}_{i=1,...,|Z_u|}
\end{math}. An estimator of mutual information, in the homogenous case is give by:
\begin{equation}
    \hat{I}(z_u,y_v)=\frac{N}{2} \log \left(\frac{s_0}{s} \right),
\end{equation}
where \begin{math}
s_0=\sum_{k=1}^N(y_v^{(k)}-\hat{y}_v)/N
\end{math} and \begin{math}
s=\sum_{i=1}^{|Z_u|}n_is_i/N
\end{math}. \begin{math}
k_{z_u,y_v}= |Z_u|-1
\end{math} are the degree of freedom associated to the mutual information in the homogenous case.\\
While,i n the heterogeneous case an estimator of the mutual information is equal to \begin{equation}
    \hat{I}(z_u,y_v)=\frac{N}{2} \log (s_0) -\frac{1}{2} \sum_{i=1,...,|Z_s|} n_i \log (s_i)
    \label{mutal}
\end{equation}
with \begin{math}
k_{z_u,y_v}= 2(|Z_u|-1) 
\end{math} degrees of freedom.
According \cite{edwards2010selecting} it is useful to use either \begin{math}
\hat{I}^{AIC}=\hat{I}(x_i,x_j)-2k_{x_i,x_j}
\end{math} or \begin{math}
\hat{I}^{BIC}=\hat{I}(x_i,x_j)-\log(n) k_{x_i,x_j}
\end{math}, where \begin{math}
k_{x_i,x_j}
\end{math} are the degree of freedom, to avoid inclusion of links not supported by the data. This aspect is suggested by the algorithm to find the best spanning tree, because it stop when it has added the maximum number of edges. Furthermore the algorithm avoid inside the tree a forbidden path.  The definition of forbidden path is a path between tow not adjacent discrete nodes which passes through continuous nodes \citep{de2009high}. However, we can start from the best spanning tree and determine the best strongly decomposable graphical model. A strongly decomposable graphical model whose graph neither contains cycles of length more than three nor forbidden path. Strongly decomposable model is an important class of model that can be used to analyze mixed data. This class restrict the class of possible interaction model which would be to huge to be explored \citep{abbruzzo2015inferring}.
The graph build to find the best spanning tree, can be see with a symmetric adjacency matrix $AM$, with dimension $V \times V$, in which each element takes value of 1 if an edge exists between two of the $V$ variables, and zero otherwise. Elements in the main diagonal are zeros, since self-loops are not allowed.

\section{A measure of dynamic stability as proxy for the model drift}
Considering the additional dimension of time $t$ to the dataset of $N$ observations and $p$ variables as a tensor $X$ with dimension \begin{math}( N \times p \times T )
\end{math},  we are interested in modeling the evolution of the joint probability  $P(X_1,...,X_p)$ over  T time periods. In other words, considering the graph $G$, with $V=p$ vertices  of the maximum spanning tree with mutual information as express in  Eq. \ref{mutal} for each period $t=1,...,T$ and the corresponding $T$ adjacency matrices $AM_{t}$, the aim of the paper is to describe how the graphs, as represented by their adjacency matrix $AM_t$ with $t=1,2,...T$ , change over time. 

\subsection{Transition Matrix Processes}

In order to accomplish this task, we analyse the transition process which connects the original adjacency matrix $AM_{1}$ to any adjacency matrices in a subsequent period $AM_{T}$. 
We first introduce a function which maps any possible state of $AM_{t}$ into a transition matrix $TM=f(AM_{t})$ with $t=1,2,3,..T$, noted $TM_{T}$, of dimension $V \times V$. Its generic element $w_{i,j}$ registers all possible states of dependence of any couple of variable $V_{i}$ and $V_{j}$ in $T$ periods. Specifically, the function takes the following form:
\begin{equation}
\label{eq:main}
\centering
   TM_{t}=\sum_{t=1}^T 2^{(T-t)} AM_{t}
\end{equation}

For the sake of clarity, the following paragraph describes the process up to $T=3$ and, thereafter, generalizes for $T$ periods.

\begin{table}[H]
\centering
\begin{tabular}{l|l|l|l|l}
\hline
                    & $AM_{1}$ & $AM_{2}$                     & $AM_{3}$ & $TM_{3}$ \\ \hline
 & 0  & \multicolumn{1}{l|}{0} & 0  & 0  \\
                    & 1  & \multicolumn{1}{l|}{0} & 0  & 4  \\
                    & 1  & \multicolumn{1}{l|}{1} & 0  & 6   \\
                    & 1  & \multicolumn{1}{l|}{1} & 1  &  7  \\
                    & 0  & \multicolumn{1}{l|}{1} & 0  &   2 \\
                    & 0  & \multicolumn{1}{l|}{0} & 1  & 1   \\
                    & 1  & \multicolumn{1}{l|}{0} & 1  &   5 \\
                    & 0  &
                    \multicolumn{0}{l|}{1} & 1  &   3
\end{tabular}
\caption{All possible $AM_t$ values for two nodes $i$ and $j$ and the resulting $w_{i,j}$ in $TM_{T}$ function for $T=3$}
\label{tab:example}
\end{table}

As a starting point, in $t=1$ the transition matrix $TM_{1}$ is equal to the adjacency matrix $AM_{t}$, where $w_{i,j;1}=0$ means that the \textit{i}-node and \textit{j}-node are not connected, while  when $w_{i,j;1}=1$  means that the \textit{i}-node and \textit{j}-node are connected. 
At $t=2$ existing links can persist or not, while non-existing links can appears or not. From Eq. \ref{eq:main}, 
\begin{equation}
\centering
TM_{2}= 2 \times AM_{1}+AM_{2}
\end{equation}
Thus, $TM_{2}$ maps any possible evolution of connections ${w}_{i,j;2}$ with values $\{0,1,2,3\}$. When $V_{i}$ and $V_{j}$ are never connected,that is $AM_{i,j;t=1}=AM_{i,j;t=2}=0$, then $TM_{i,j;2}=0$. If $V_{i}$ and $V_{j}$ stay connected, that is $AM_{i,j;t=1}=AM_{i,j;t=2}=1$, then $w_{i,j;2}=3$. For $AM_{i,j}$ changing from 0 in $t=1$ to 1 in $t=0$ and viceversa, we have ${w}_{i,j;2}=2$ and ${w}_{i,j;2}=1$, respectively. At time $t=3$ the possible evolution of $AM$ can be described has 8 levels, since it can be either 0 or one three times, given by:

\begin{equation}
\centering
TM_{3}=2^2\times AM_{1}+2^1\times AM_{2}\textit{}+ 2^0\times AM_{3} 
\end{equation}

Table \ref{tab:example} summarizes all possible combinations between two nodes of binary values of the $AM_{t}$ in the three periods, mapped on $TM_{3}$. Generally, for time $T$ we can derive Eq. \ref{eq:main}:

\begin{equation}
\centering
\begin{split}
TM_{2}= 2 \times AM_{1}+AM_{2}\\
TM_{3}= 2 \times TM_{1,2}+AM_{3}\\
TM_{3}= 2 \times ( 2 \times AM_{1}+AM_{2})+AM_{3}\\
TM_{3}=2^2\times AM_{1}+2^1\times AM_{2}+ 2^0\times AM_{3} \\
TM_{3}=\sum_{t=1}^T 2^{(3-t)} AM_{t}\\
...\\
TM_{T}=\sum_{t=1}^T 2^{(T-t)} AM_{t}
\end{split}
\end{equation}

In general, the value of the generic element \begin{math}
w_{i,j;t} \in  \mathcal{W} \subset \mathbb{N}
\end{math} of $TM_{t}$ can be considered as a discrete random variable with density
$f(\textit{w}_{i,j;t})$

\begin{equation}
\centering
    f({w}_{i,j;t})= P(\mathcal{W}_{i,j;t}=\textit{w}_{i,j;t}) , \quad {t}=2,...,T
\end{equation}

Thus, $w_{i,j;t}$ represents the evolution of the connection between \textit{i}-node with \textit{j}-node at time $T$, for each node \textit{V}. The numerosity of the set $\mathcal{W}_{i,j;T}=\{0,1,2..,2^T -1\}$ is $2^T$. 

\subsection{From the transition process to stability}
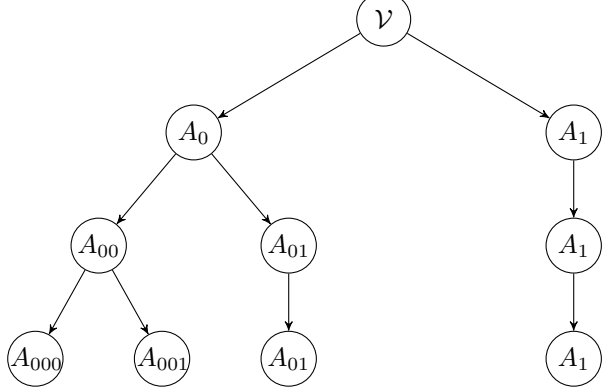
\begin{figure}[H]
     \centering
     \textbf{Partition of Transition Matrix Process}
     \tikzset{
  treenode/.style = {align=center, inner sep=0pt, text centered,
    font=\sffamily},
  arn_n/.style = {treenode, circle, black, font=\sffamily\bfseries, draw=black,
    fill=white, text width=1.9em}
}

\begin{tikzpicture}[->,>=stealth',level/.style={sibling distance = 5cm/#1,
  level distance = 1.5cm}] 
\node [arn_n] {\begin{math}\mathcal{V}\end{math}}
    child{ node [arn_n] {\begin{math}A_0\end{math}
    } 
            child{ node [arn_n] {\begin{math}A_{00}\end{math}}
            child{node[arn_n]
            {\begin{math}A_{000}\end{math}
    }}
            child{node[arn_n]
            {\begin{math}A_{001}\end{math}
    }}
            	}
            child{ node [arn_n] {\begin{math}A_{01}\end{math}}
            child{node[arn_n]
            {\begin{math}A_{01}\end{math}
    }}
            }                            
    }
    child{ node [arn_n] {\begin{math}A_{1}\end{math}}
            child{ node [arn_n] {\begin{math}A_{1}\end{math}} 
			child{ node [arn_n] {\begin{math}A_{1}\end{math}} 
							}}
		}
; 
\end{tikzpicture}
     \caption{Representation of \textit{Transition Matrix process} with Tail-free processes}
     \label{fig:2}
 \end{figure}

The main idea of the paper is to consider as a proxy for the model drift the appearance or disappearance of connections between nodes, that is changes of the conditional independence structure of a dataset over time. For this reason, we are specifically interested in two specific levels. The one describing the state of the word in which a connection between two nodes never exists, that is $AM_{i,j;t}=0$ $\forall$ $t$ and the one describing a stable connection over time, that is $AM_{i,j;t}=1$ $\forall$ $t$. For the case $T=3$, the two cases map into $w_{i,j;3}=0$ and $w_{i,j;3}=7$, as showed in Table \ref{tab:example}. In general for a generic $T$, we have a stability of connections when connections are always absent, with $w_{i,j;T}=0$, or always existing, with $w_{i,j;T}=2^{T}-1$. This transition process is a partition process (Fig \ref{fig:2}) of the set of  $\mathcal{V}$   possible connections between the $\textbf{V}$ nodes in the undirected graph: $\mathcal{V}=\frac{V(V-1)}{2}$. Each transition in time $t$ generates a subsequent partition of $\mathcal{V}$, one of whose will always contain elements for which  $w_{i,j;t}=0$ or always $w_{i,j;t}=2^{t}-1$.  This \textit{transition processes} is a special case of the Tail-free processes \citep{jara2011class}. 
Consider a sequence $\mathcal{T}_0=\{\mathcal{V}\}$, $\mathcal{T}_1=\{A_0,A_1\}$, $\mathcal{T}_2=\{A_{00},A_{01},A_{1}\} $, and so on, of measurable partitions of the \begin{math}
\mathcal{V}
\end{math}  elements, obtained by slitting every set in the preceding partition into two new sets for the node on left and maintain the same node for the others. 
Specifically, at each time $t$ we can partition the elements between stable and unstables ones. Fig. \ref{fig:2}  shows a tree diagram that represents the distribution of mass over time \begin{math}
 \mathcal{V}= A_0 \cup A_1= (A_{00} \cup A_{01}) \cup A_{10}
\end{math} of the elements at each time. 
$A_{0}$ contains elements for $w_{i,j,2}={0,3}$, that is stable connections while $A_{1}$, the remaining ones. At the subsequent period, $A_{0}$ is partitioned between $A_{00}$, in which connection remain stable with $w_{i,j,3}={0,7}$,while  $A_{01}={1,6}$ and $A_{1}$ the remaining ones.

Clearly, every partition is composed by the union of all possible evolution of the connection given by the levels of  $\mathcal{W}$, and, by construction, there is always a partition with elements $w_{i,j;t}=0$ and $w_{i,j;t}=2^{t}-1$, that containing stable links  between the \textit{i}-node and  the \textit{j}-node until time $t$. 
We describe this process as a variable $Y_{i,j;t}$ with values : 
\begin{equation}
\label{eq:y}
    {Y}_{i,j;t} = 
    \begin{cases}
      y_{i,j;t}=1 & \text{if   } w_{i,j;t}=0 \lor w_{i,j;t}=2^{t}-1 \\
      y_{i,j;t}=0 & \text{otherwise}
    \end{cases}, 
    \quad
    t=2,...,T
\end{equation}
Thus, $Y_{i,j;t}$ is indicate persistent status of dependence over time \begin{math}
Y_{i,j;k}=1
\end{math} or not \begin{math}
Y_{i,j;k}=0
\end{math}. 
Be $Y_t$ the vectorization of $Y_{i,j;t}$, $vec(Y_{i,j;t})=Y_t$ with length $\mathcal{V}=\frac{V \times (V-1)}{2}$, that is at each time we observe the stability of the $\mathcal{V}$ connection between each possible pair of nodes.   The structure of the \textit{transition matrix process}  depend by the spanning forest at time \begin{math}
t=1
\end{math}, and for each period we have a partition of $\mathcal{V}$ given by \begin{math}
\mu_t=\sum_{i=1}^N Y_{i,t}
\end{math} with \begin{math}
t=1,...,T-1
\end{math}.

Therefore, we pool together the $T-1$ periods and define $Stability$, the resulting variable $Y$ with length $n=\mathcal{V} \times (T-1)$. $Stability$ is the cornerstone of our strategy to estimate an empirical measure of model drift.

\subsection{The stability index}

In this section we introduce the \textit{Stability} as a latent variable, which capture stability of connection of a graph overtime.\\ 
Consider the following variable with same length $i=1,...,n$:
\begin{itemize}
    \item $Y$, $Stability$ as defined above
    \item $W=vec(TM_{i,j,t})$ that is the vectorization of the value $w_{i,j;t}$ of $TM$.
    \item $T$ the corresponding time for each $Y_i$.
\end{itemize}
We build a dataset with this variables and call it $\textbf{D}$.
Note that by construction the observations of $\textbf{D}$ is exchangeable since we have built $\textbf{D}$ respecting the temporal period of the \textit{adjacent matrices}, thus:
\begin{equation*}
    \centering
    P(\textbf{D}_1,...,\textbf{D}_n)=P(\textbf{D}_{\sigma(1)},...,\textbf{D}_{\sigma(n)})
\end{equation*}
for all \begin{math}
n \geq 1
\end{math} and all permutations \begin{math}
\sigma 
\end{math} of \begin{math}
(1,...,n)
\end{math}. In other words, the order of appearance of the observation does not matter in terms of their joint distribution. Let  $\theta_i$ the probability of a realization of $Y_i=1$ of \textit{Stability} with odds of stability $\frac{\theta_i}{1-\theta_i}$. Thus
the dichotomous variable $Y$ can be described by a Bernoulli distribution with probability of success $
\theta_i$:
\begin{equation*}
    Y_i|\theta_i \overset{ind}{\sim} Bern(\theta_i), \quad i=1,..,n
\end{equation*}

Consider a logistic regression model\footnote{The logistic regression seem the most natural way to describe this phenomenon. However, according to the type of expected drift, we could employ other function, without loss of generalization.}, which writes that the logit of the probability $\theta_i$, or the log of the its odd is a linear function of some predictor variables $\textbf{x}_i$:
\begin{equation}
\text{Logit}(\theta_i)=\log \left(\frac{\theta_i}{1-\theta_i} \right)= \beta_0+ \sum_{j}^{2^t} \beta_j \textbf{x}_{j,i}
\label{15}
\end{equation}

where the $j$ predictors are $T$, that is the time of the realization of $Y$ and $W$, that is the corresponding value. Since $W$ has $2^t$ levels, we regress $2^t-1$ dummy variable and keep $W={0}$ as the reference category:

\begin{equation}
\log \left(\frac{\theta_i}{1-\theta_i} \right)= \beta_0+ \beta_1 \times T +\sum_{j}^{2^t-1} \beta_j \textbf{w}_{j,i}
\label{eq:16}
\end{equation}

By construction, the intercept of this model $\beta_0$ can be interpreted as the baseline risk for $Stability$. A high $\beta_0$ suggests that the underlying graphical model is not changing much over time. $\beta_t$ captures the effect of the drift over time. It can be shown that $Stability$ is weakly decreasing over time and, thus $\beta_1$ define the speed of convergence towards the absence of stability. Finally, since the variable $Y$ takes value 1 for $W_{i,j}=(0,T^2-1)$, the coefficient $\beta_{T^2-2}$, that is the coefficient for $W_{i,j}=T^2-1$ with reference $W_{i,j}=0$ captures which component of $Stability$ originates in the persistence of existing connections, rather than on the persistence of absence of connections. 

The computation is straightforward: by rearranging the logistic regression Equation \ref{15}, it is possible to express the regression as a nonlinear equation for the probability of success \begin{math}
\theta_i:
\end{math}
\begin{equation}
\begin{split}
    \log \left(\frac{\theta_i}{1-\theta_i} \right)= \beta_0+\sum_{j}^p \beta_j \textbf{x}_{j,i} \\
 \frac{\theta_i}{1-\theta_i}=\exp \left\{ \beta_0+\sum_{j}^p \beta_j \textbf{x}_{j,i} \right\}\\
\theta_i=\frac{\exp \left\{ \beta_0+\sum_{j}^p \beta_j \textbf{x}_{j,i} \right\}}{1+\exp \left\{ \beta_0+\sum_{j}^p \beta_j \textbf{x}_{j,i} \right\}}
\end{split}
\label{15bis}
\end{equation}
From the Equation \ref{15bis} we can define the likelihood for the sequence of \begin{math}
Y_i
\end{math} over data set of \textit{n} subjects is then

\begin{figure}[!htbp]
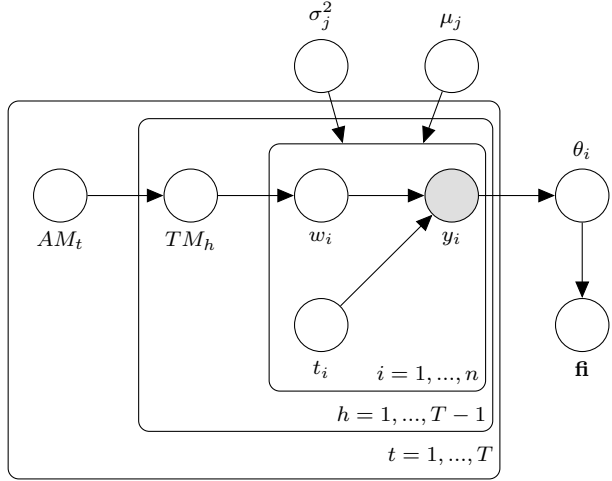

\centering
\textbf{Stability Process}\par\medskip
 \caption{Bayesian Graphical Model of the \textit{stability}}
  \tikz{ %
    \node[latent] (AM) [label=below:\begin{math}AM_t\end{math}]{} ; %
    \node[latent, right=of AM] (TM) [label=below:\begin{math}TM_h\end{math}]{} ; %
    \node[latent, right=of TM] (w)[label=below:\begin{math}w_i\end{math}]{} ; %
    \node[latent, below=of w] (t) [label=below:\begin{math}t_i\end{math}]{} ; %
    \node[obs, right=of w] (y) [label=below:\begin{math}y_i\end{math}]{} ; %
    \node[latent, above=of y] (b) [label=above:\begin{math}\mu_j\end{math}]{};%
    \node[latent, left=of b] (s) [label=above:\begin{math}\sigma_j^2\end{math}]{};%
    \node[latent, right=of y] (theta) [label=above:\begin{math}\theta_i\end{math}]{};%
    \node[latent, below=of theta] (beta) [label=below:\begin{math}\mathbf{\beta}\end{math}]{};%
    \plate[inner sep=0.20cm, xshift=-0.12cm, yshift=0.12cm] {plate 1} {(w) (y) (t)} {\begin{math}i=1,...,n\end{math}};%
    \plate[inner sep=0.20cm, xshift=-0.12cm, yshift=0.12cm] {plate2} {(TM) (plate 1)} {\begin{math}h=1,...,T-1\end{math}}; %
    \plate[inner sep=0.20cm, xshift=-0.12cm, yshift=0.02cm] {plate3} {(AM) (plate2) (plate 1)} {\begin{math}t=1,...,T\end{math}}; %
    \edge {AM} {TM} ; %
    \edge {TM} {w} ; %
    \edge {w,t} {y} ; %
    \edge{s,b} {plate 1} ; %
    \edge{y} {theta} ; %
    \edge{theta} {beta} ; %
    }
    \label{BG}
\end{figure}

\begin{dmath}
p(\textbf{D}|\beta_0,\boldsymbol\beta_p )=
\prod_{i=1}^{n}\left[
\left(\frac{\exp \left\{ \beta_0+\sum_{j}^p \beta_j \textbf{x}_{j,i} \right\}}{1+\exp \left\{ \beta_0+\sum_{j}^p \beta_j \textbf{x}_{j,i} \right\}}\right)^{y_i}
\\
\left(1-\frac{\exp \left\{ \beta_0+\sum_{j}^p \beta_j \textbf{x}_{j,i} \right\}}{1+\exp \left\{ \beta_0+\sum_{j}^p \beta_j \textbf{x}_{j,i} \right\}}\right)^{(1-y_i)}\right]
\label{16}
\end{dmath}

where $\textbf{D}$ is the dataset composed by $T_i$ and the corresponding dummy variables generated by the level of $W_i$. The set of unknown parameters consists of \begin{math}
\beta_0,\beta_T,...,\beta_{T^2-2} \end{math}. In general, any prior distribution can be used, depending on the available prior information. The literature  suggests  the  use  of  informative  prior  distributions  if  something  is known about the likely values of the unknown parameters, otherwise, the use of non-informative prior if either little is known about the coefficient values or if one wishes to see what the data themselves provide as inferences. In this case, we will use the most common priors for logistic regression parameters:

\begin{equation}
    \beta_j \sim N(\mu_j,\sigma_j^2)
    \label{17}
\end{equation}
The most common choice for $\mu$ is zero with $\sigma$ large enough to considered as non-informative in the range from $\sigma=10$ to $\sigma=100$. The posterior distribution of \begin{math}
\boldsymbol \beta_j
\end{math} is extrapolated by combining likelihood Eq. \ref{16}, with the prior in Eq. \ref{17}:

\begin{dmath}
p(\beta_0,\boldsymbol\beta_p|\textbf{D},\sigma_j, \mu_j)=
\prod_{i=1}^{n}\left[
\left(\frac{\exp \left\{ \beta_0+\sum_{j}^p \beta_j \textbf{x}_{j,i} \right\}}{1+\exp \left\{ \beta_0+\sum_{j}^p \beta_j \textbf{x}_{j,i} \right\}}\right)^{y_i}\\
\left(1-\frac{\exp \left\{ \beta_0+\sum_{j}^p \beta_j \textbf{x}_{j,i} \right\}}{1+\exp \left\{ \beta_0+\sum_{j}^p \beta_j \textbf{x}_{j,i} \right\}}\right)^{(1-y_i)}\right]
\times \prod_{j=0}^p\frac{1}{\sqrt{2\pi}\sigma_j}\exp \left\{ -\frac{1}{2}\left(\frac{\beta_j-\mu_j}{\sigma_j}\right)^2\right\}
\label{18}
\end{dmath}
Now, we are not that much interested in the regression parameters \begin{math}
\boldsymbol \beta_j
\end{math}, we want to find the posterior probability distribution of the \textit{stability}. Furthermore, this model gives us the opportunity to compute the prediction of the \textit{stability} over a specific time \textit{t}. If \begin{math}
\widetilde{y_i}
\end{math} represents the number of similarity connection between \textit{n} nodes at time \begin{math}
t
\end{math}, then one would be interested in the posterior predictive distribution of the fraction \begin{math}
\widetilde{y_i}/n
\end{math} One represents this predictive density of \begin{math}
\widetilde{y_i}
\end{math} as:
\begin{equation}
    f(\widetilde{Y_i}|y)=\int{p(\beta_0,\boldsymbol\beta_p|\textbf{D},\sigma_j, \mu_j)p(\widetilde{y_i}, \textbf{X}|\beta_0,\boldsymbol\beta_p ) d\boldsymbol\beta}
\end{equation}
where \begin{math}
p(\beta_0,\boldsymbol\beta_p|\textbf{D},\sigma_j, \mu_j)
\end{math} is the posterior density of \begin{math}
\boldsymbol\beta
\end{math} and \begin{math}
p(\widetilde{y_i}, \textbf{X}|\beta_0,\boldsymbol\beta_p )
\end{math}  is the Binomial sampling density of \begin{math}
\widetilde{y_i}
\end{math} conditional of regression vector \begin{math}
\boldsymbol\beta=(\beta_0,\boldsymbol\beta_p )
\end{math}. Figure \ref{BG} represents the Bayesian graphical model of the stability, in particular, we can see all process that describes from the adjacent matrix to the coefficients of the logistic, that say us how changes the relationship between the variables over the time. Where we have an adjacent matrix \begin{math}
(AM)
\end{math} for each time \begin{math}
t
\end{math}, for each pair sequential of the \textit{AM} we have a transition matrix \begin{math}
TM
\end{math}. From the \textit{TM} we can build the dataset to compute the stability with \textit{n} observation, where \begin{math}
n=\mathcal{V}\times (T-1)
\end{math}, and three  variables: $\textbf{W,T,Y}$. 

\section{Empirical experiment}
As a test bed for this theoretical approach, we apply the stability index to the \textit{ELEC2} dataset \citep{harries1999splice}, a benchmark for  drift evaluation \citep[among the many]{baena2006early,kuncheva2008adaptive}. It holds information on the Australian New South Wales (NSW) Electricity Market, containing $27552$ records dated from May 1996 to December 1998, each referring to a period of 30 minutes. These records have 5 fields: a binary class label $Y$ and four covariates $X_1, X_2, X_3$ and $X_4$ capturing different aspects of electricity demand and supply. In order to compute the empirical evolution of the drift over time, we group observations in one week period. Thus, for each week we have a panel dataset of 5 variables and 336 observation. Thus, we have a tensor $X$ with dimension $(N\times p \times T)$ with $N=336$ records for a week, $p=5$ the variables as described above and  $T=82$ temporal periods.

\begin{figure}[!htbp]

\begin{tabular}{cc}
\textbf{Spanning Tree $t=1$} & \textbf{Spanning Tree $t=8$}\\
\includegraphics[width=0.20\textheight]{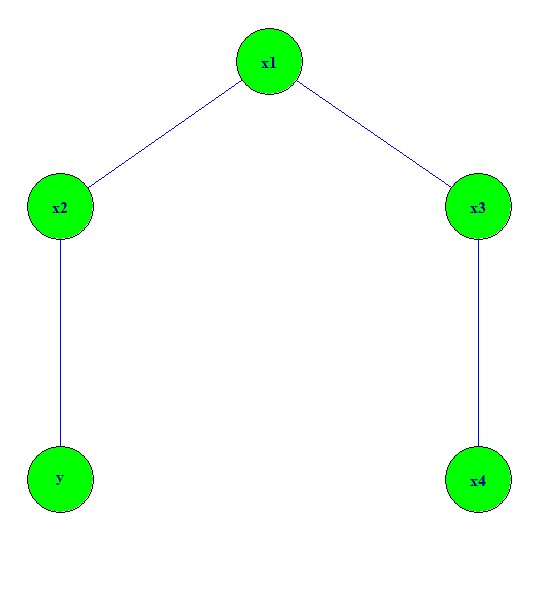} &
\includegraphics[width=0.20\textheight]{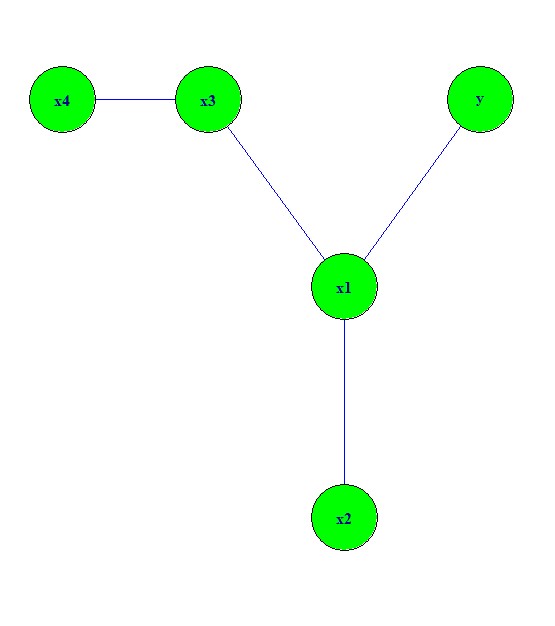}\\
\textbf{Spanning Tree $t=12$} & \textbf{Spanning Tree $t=14$}\\
\includegraphics[width=0.20\textheight]{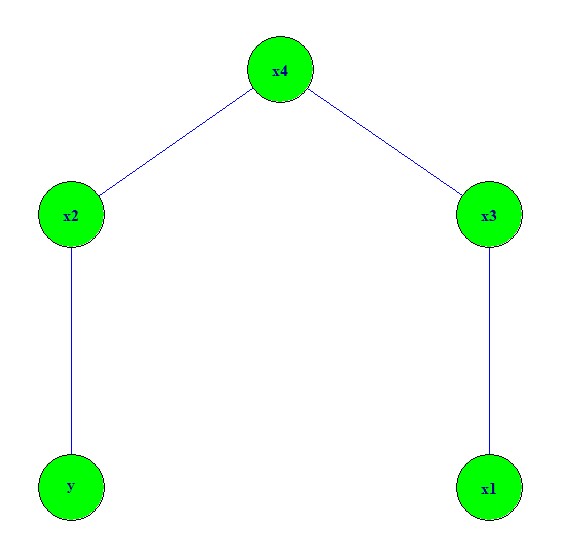}&
\includegraphics[width=0.20\textheight]{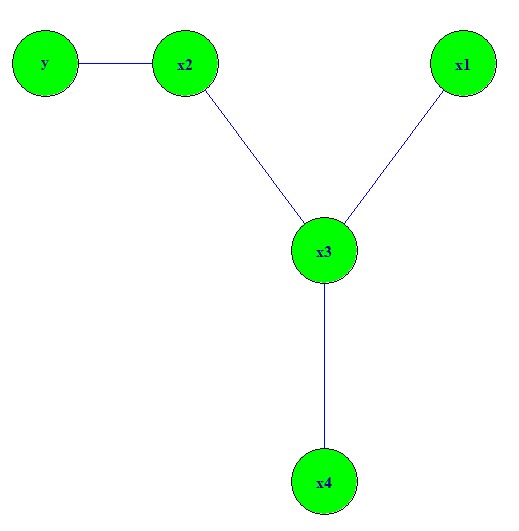}\\
\textbf{Spanning Tree $t=19$} & \textbf{Spanning Tree $t=41$}\\
\includegraphics[width=0.20\textheight]{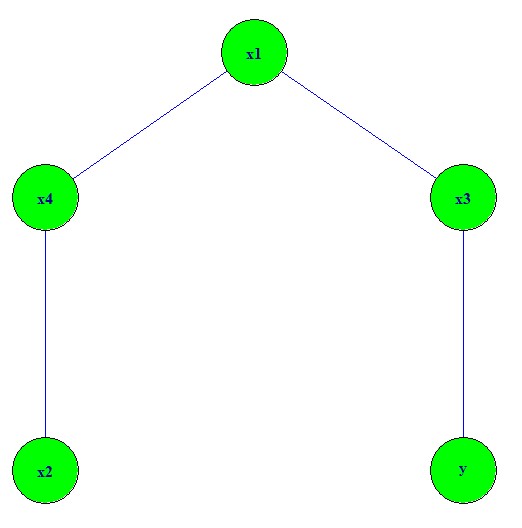} &
\includegraphics[width=0.20\textheight]{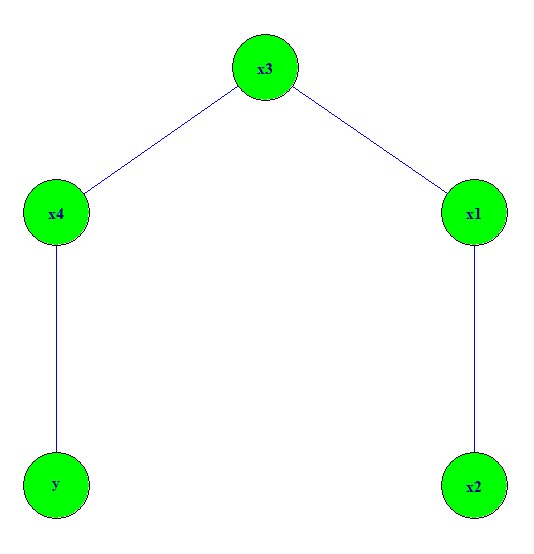}\\
\end{tabular}
\caption{Graph over the time}
 \label{DriftMercato}
\end{figure}

First, we realize a  Graphical Models for each period  $t$ as the start point of our strategy to compute the drift. Figure \ref{DriftMercato} portraits the graphs for some selected periods and shows that the structure of the graph changes overtime. We thus expect  a presence of the drift. 

\begin{figure}
\centering
\par\medskip
\includegraphics[scale=0.58]{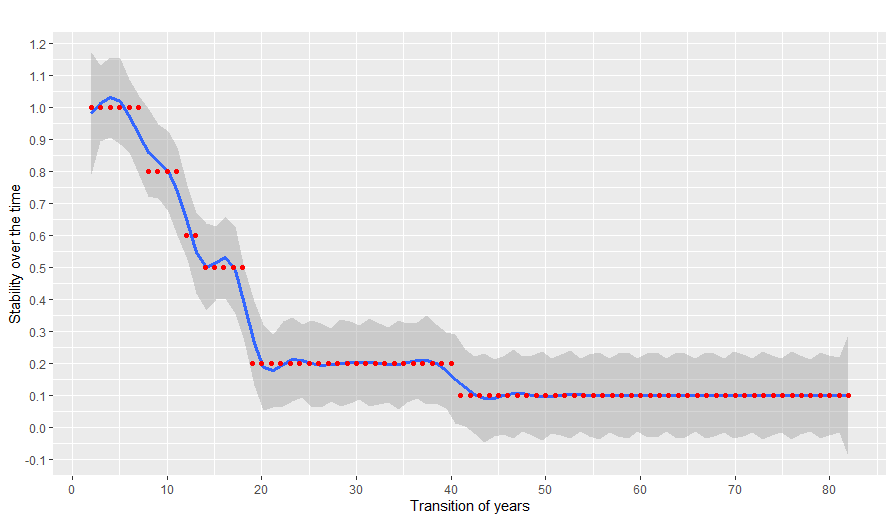}
\caption{Evolution of Stability}
\label{logit}
\end{figure}

Figure \ref{logit} depicts the evaluation of the drift overtime. The red dots are the percentage of stable relations among variables, that is the the sum of variable $Y_{i,t}$ in Equation \ref{eq:y}, while the blue line is the estimation of the Equation \ref{18} with its related confidence interval as the gray contour.The figure highlights $6$ periods of drift. The different \textit{Stability} values are reported in the table \ref{tab2}. In the table \ref{tab3} are reported 
the magnitude of the coefficients for the baseline $\beta_0$ or intercept, $\beta_{2^T-1}$ for the $W=2^T-1$ with reference level $W=0$  and for the time $\beta_{time}$.

\begin{table}[!htbp]
\label{tab2}
\caption{Approximation of the drift for selected period }\par\medskip
\begin{tabular}{|c|l|l|l|l|l|l|}
\hline
\multirow{2}{*}{\textbf{\begin{tabular}[c]{@{}c@{}}Percent \\ of Stability\end{tabular}}} & \multicolumn{6}{c|}{\textbf{Evolution of the Drift}}                                                                                                                                                                                                                                                                                                                                               \\ \cline{2-7} 
                                                                                          & {\begin{tabular}[c]{@{}l@{}}$ty=2$\end{tabular}} & {\begin{tabular}[c]{@{}l@{}}$ty=8$\end{tabular}} & {\begin{tabular}[c]{@{}l@{}}$ty=12$\end{tabular}} & {\begin{tabular}[c]{@{}l@{}}$ty=14$\end{tabular}} & {\begin{tabular}[c]{@{}l@{}}$ty=19$\end{tabular}} & {\begin{tabular}[c]{@{}l@{}}$ty=41$\end{tabular}} \\ \hline
\textbf{$ \frac{\sum_{i=1}^N Y_{i,t}}{N}$}                                                & \textbf{$1.0$}                                                  & \textbf{$0.8$}                                                  & \textbf{$0.6$}                                                & \textbf{$0.5$}                                                    & \textbf{$0.2$}                                                   & \textbf{$0.1$}                                                    \\ \hline
\end{tabular}
\end{table}


\begin{table}[!htbp]
\small
\textbf{Regression Summary}\par\medskip
\begin{tabular}{@{}|l|l|@{}}

\textbf{Coefficients}    & \textbf{Estimation} \\ 
\textbf{$\beta_0$}       & \textbf{7.66}       \\ 
\textbf{$\beta_{2^T-1}$} & \textbf{19.75}      \\ 
\textbf{$\beta_{Time}$}  & \textbf{-0.30}      \\ 
\end{tabular}
\caption{Coefficients of logistic regression }
\label{tab3}
\end{table}

\section{Conclusion}

This paper presented an algorithm to estimate the magnitude of a model drift in a context of machine learning. While past solutions relies on how the classification errors of a specific target variable changes over time, the present method tries to describe the underlying hidden context with the use of graphical models and to estimate how the observable context changes over time. Specifically, we provide not only an assessment of the drift, which is independent from the model in use, but also an estimation of the confidence interval of this prediction. These two characteristics combined together allow to signal when a data driven process shows an excessive risk due to the drift and needs to be retrained or re-calibrated. Possible applications are countless such as predicting defaults, online recommendations systems, or spam filtering. More specific, any prediction which involves human behaviour is prone to constant changes in the data generating process, while biological and physical phenomena tend to be more stable over time. Further lines of research in this area include a fine tuning for estimating different type of drift, allowing for temporary drift, and testing the index on a wider array of applications.

\bibliography{literature}
\end{document}